\documentclass[letterpaper]{article} 
\usepackage{aaai24}  
\usepackage{times}  
\usepackage{helvet}  
\usepackage{courier}  
\usepackage[hyphens]{url}  
\usepackage{graphicx} 
\usepackage{natbib}  
\usepackage{caption} 
\usepackage[colorlinks,
            linkcolor=black,
            anchorcolor=black,
            citecolor=black]{hyperref}

\usepackage{listings}
\usepackage{xcolor}
\usepackage[switch]{lineno} 
\usepackage{amssymb}
\usepackage{amsmath}
\usepackage{multirow}
\usepackage{booktabs}
\usepackage[ruled,linesnumbered]{algorithm2e}
\SetKwInput{KwInput}{Input}                %
\SetKwInput{KwOutput}{Output}              %

\newcommand{\etal}{et al.}
\newcommand{\Tref}[1]{Table~\ref{#1}}
\newcommand{\Eref}[1]{Eq.~(\ref{#1})}
\newcommand{\Fref}[1]{Fig.~\ref{#1}}

\newcommand{\Aref}[1]{Algorithm.~\ref{#1}}

\newcommand{\ie}{i.e.}
\newcommand{\eg}{e.g.}

\title{Data-Free Hard-Label Robustness Stealing Attack}
\author{
    Xiaojian Yuan\textsuperscript{\rm 1},
    Kejiang Chen\footnote{Corresponding author.}\textsuperscript{\rm 1},
    Wen Huang\textsuperscript{\rm 1}, 
    Jie Zhang\textsuperscript{\rm 2}, \\
    Weiming Zhang\textsuperscript{\rm 1}, 
    Nenghai Yu\textsuperscript{\rm 1}
}
\affiliations{
    \textsuperscript{\rm 1}University of Science and Technology of China, 
    \textsuperscript{\rm 2}Nanyang Technological University \\
    
    xjyuan@mail.ustc.edu.cn, chenkj@ustc.edu.cn, hw2000@mail.ustc.edu.cn, \\ jie\_zhang@ntu.edu.sg, zhangwm@ustc.edu.cn, 
    ynh@ustc.edu.cn

}

\usepackage{bibentry}

\begin{document}

\maketitle

\begin{abstract}
The popularity of Machine Learning as a Service (MLaaS) has led to increased concerns about Model Stealing Attacks (MSA), which aim to craft a clone model by querying MLaaS. Currently, most research on MSA assumes that MLaaS can provide soft labels and that the attacker has a proxy dataset with a similar distribution. However, this fails to encapsulate the more practical scenario where only hard labels are returned by MLaaS and the data distribution remains elusive. Furthermore, most existing work focuses solely on stealing the model accuracy, neglecting the model robustness, while robustness is essential in security-sensitive scenarios, \eg, face-scan payment. Notably, improving model robustness often necessitates the use of expensive techniques such as adversarial training, thereby further making stealing robustness a more lucrative prospect. In response to these identified gaps, we introduce a novel Data-Free Hard-Label Robustness Stealing (DFHL-RS) attack in this paper, which enables the stealing of both model accuracy and robustness by simply querying hard labels of the target model without the help of any natural data. Comprehensive experiments demonstrate the effectiveness of our method. The clone model achieves a clean accuracy of 77.86\% and a robust accuracy of 39.51\% against AutoAttack, which are only 4.71\% and 8.40\% lower than the target model on the CIFAR-10 dataset, significantly exceeding the baselines. Our code is available at: \textit{https://github.com/LetheSec/DFHL-RS-Attack}.

\end{abstract}

\section{Introduction}\label{sec:intro}
Machine learning as a service (MLaaS) has gained significant popularity due to its ease of deployment and cost-effectiveness, which provides users with pre-trained models and APIs. Unfortunately, MLaaS is susceptible to privacy attacks, with Model Stealing Attacks (MSA) being particularly harmful~\cite{tramer2016stealing,orekondy2019knockoff,jagielski2020high,yuan2022attack,wang2022black}, where an attacker can train a clone model by querying its public API, without accessing its parameters or training data. 
This attack not only poses a threat to intellectual property but also compromises the privacy of individuals whose data was used to train the original model. Moreover, the clone model can serve as a surrogate model for other black-box attacks, \eg, adversarial examples (AE)~\cite{, Zhang_2022_CVPR}, membership inference~\cite{shokri2017membership}, and model inversion~\cite{yuan2023pseudo}.

Furthermore, numerous security-sensitive scenarios require deployed models are not only accurate but also robust to various attacks, such as adversarial attacks. To address this issue, MLaaS providers can employ adversarial training (AT) techniques~\cite{madry2017towards} to improve the robustness of their models~\cite{goodman2020attacking,shafique2020robust}. Despite the target model's robustness, most existing MSA are limited to Accuracy Stealing, \ie, reconstructing a model with similar accuracy to the target model, and fail at Robustness Stealing, \ie, acquiring the adversarial robustness of the target model while maintaining accuracy. Since the improvement of model robustness requires much more computational resources and extra data~\cite{schmidt2018adversarially,gowal2021improving}, robustness stealing will bring greater losses to MLaaS providers. Moreover, if an attacker seeks to train a clone model for transfer-based adversarial attacks against a robust target model, then it becomes crucial to employ robustness stealing to achieve effective attack performance~\cite{Dong_2020_CVPR, gao2020patch}.

In addition, most previous MSA require MLaaS to provide prediction logits, \ie, soft labels. However, this requirement is overly stringent in typical scenarios where MLaaS can only return top-1 prediction, \ie, hard label, for each query. Since models that require robustness are more likely trained on sensitive or private datasets, it is difficult for attackers to obtain public data with similar distributions, let alone access to the original data.
Hence, the investigation of MSA targeting robustness in a data-free hard-label setting is highly valuable and  remains unexplored.

To tackle the above issues, we propose Data-Free Hard-Label Robustness Stealing (DFHL-RS) attack, which can effectively steal both the accuracy and robustness of the target model. We first demonstrate that direct use of AT during MSA is suboptimal and point out the limitations of using Uncertain Example (UE)~\cite{li2023extracting} for robustness stealing. Then the concept of High-Entropy Example (HEE) is introduced, which can characterize a more complete classification boundary shape of the target model. By imitating the target model's prediction of HEE, the clone model gradually approaches its classification boundaries, so as to achieve prediction consistency for various samples. Moreover, to eliminate the reliance on natural data and the logits of the target model, we design a data-free robustness stealing framework in a hard-label setting. Specifically, we first train a generator to synthesize substitute data for approximating the distribution of the target data. Since only hard labels are available, we cannot use the target model for gradient backpropagation~\cite{chen2019data,zhang2022qekd} or gradient estimation~\cite{truong2021data,kariyappa2021maze}. Thus, we use the clone model as a surrogate to guide the direction of the synthesized images. To prevent the generator from overfitting to the clone model, we adopt label smoothing and data augmentation techniques. Then we sample multiple batches from the memory bank storing synthesized images and use the proposed algorithm to construct HEE. Finally, we employ HEE to query the target model and obtain pseudo-labels for training the clone model.

Our contributions can be summarized as follows:
\begin{itemize}
    \item For the first time, we explore a novel attack namely Data-Free Hard-Label Robustness Stealing (DFHL-RS) to achieve both accuracy and robustness stealing by leveraging only hard labels without any natural data.
    \item We propose the concept of High-Entropy Examples (HEE), which can better characterize the complete shape of the classification boundary.
    \item Extensive experiments demonstrate the effectiveness and stability of our proposed attack framework under various configurations.

\end{itemize}

\section{Related Work}\label{sec:related_work}

 \begin{figure*}
\centering
\includegraphics[width=0.95 \linewidth]{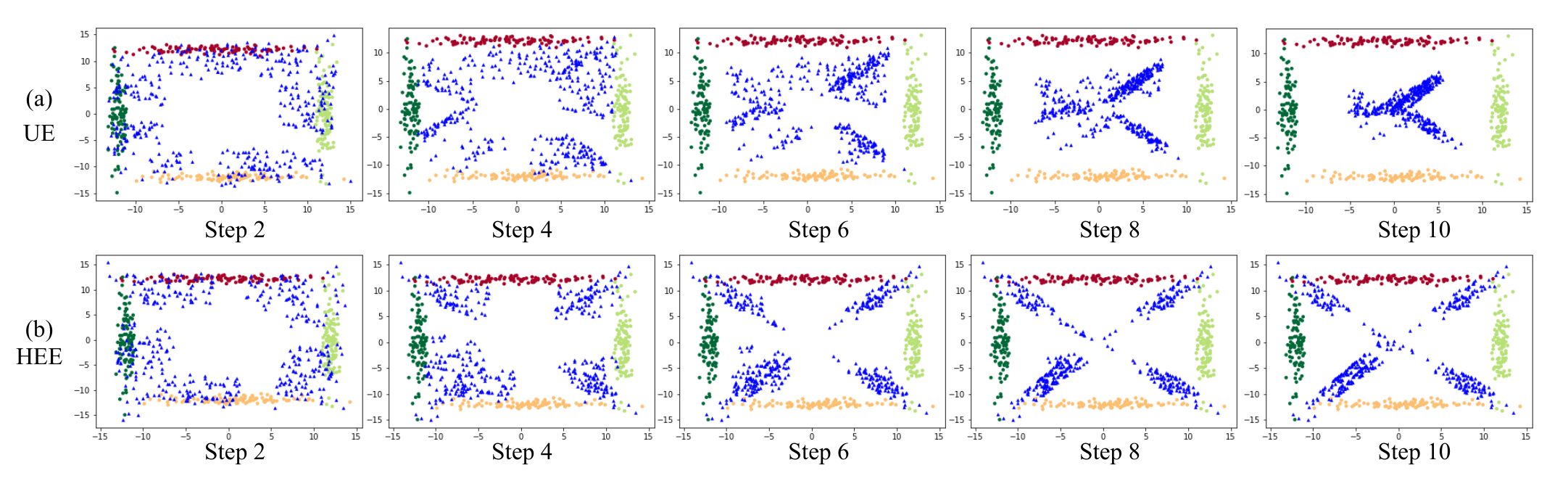}
\caption{Illustration of the superiority of HEE over UE in characterizing the classification boundaries. We make a two-dimensional dataset with four classes to train a two-layer MLP, represented by four colors. Blue points in (a) and (b) represent UE and HEE constructed at different steps according to \Eref{eq:UE} and \Eref{eq:HEE}, respectively.}
\label{fig:db_comp}
\end{figure*}

\paragraph{Data-Free Knowledge Distillation.} 
Knowledge distillation~\cite{hinton2015distilling} aims to transfer the knowledge of a large teacher model to a smaller student model. In some cases, it is not feasible to access the training data due to storage costs or privacy concerns. Therefore, some proposed distillation techniques utilizing proxy datasets with similar distributions~\cite{lopes2017data, addepalli2020degan}. ZSKD~\cite{nayak2019zero} first proposed Data-Free Knowledge Distillation (DFKD), which uses the teacher's predictions to optimize synthetic data. DAFL~\cite{chen2019data} introduced the generator for synthesizing query samples, and proposed several generative losses to promote the diversity. Adversarial DFKD~\cite{micaelli2019zero} utilized adversarial learning to explore the data space more efficiently. Some follow-up work attempted to mitigate the catastrophic overfitting ~\cite{binici2022preventing,binici2022robust}, mode collapse~\cite{ijcai2021p327} in DFKD, and to speed up the training process~
\cite{fang2022up}. However, all of these methods necessitate white-box access to the teacher. ZSDB3KD~\cite{wang2021zero} proposed DFKD in black-box scenarios, but it has high computational costs and requires a large number of queries (4000 million).

\paragraph{Data-Free Model Stealing.} 
The main difference between Data-Free Model Stealing (DFMS) and DFKD is that it only has black-box access to the teacher model, \ie, the target model. Some early work required the use of a proxy dataset for attacks~\cite{orekondy2019knockoff,barbalau2020black,wang2022black}. Based on Adversarial DFKD, recent works MAZE~\cite{kariyappa2021maze} and DFME~\cite{truong2021data} utilized gradient estimation techniques to achieve DFMS, which require the target model to return soft labels. Therefore, DFMS-HL~\cite{sanyal2022towards} extended the problem to the hard-label setting. However, it still needs to use a proxy dataset or a synthetic dataset of random shapes generated on colored backgrounds, which breaks the truly data-free setting. To address this issue, DS~\cite{beetham2023dual} proposed to train two student models simultaneously, which allows the generator to use one of the students as a proxy for the target model. However, these methods only achieved accuracy stealing, but cannot obtain  the model's robustness.

\paragraph{Adversarial Robustness Distillation.} 
Large models tend to be more robust than small models due to their greater capacity. Therefore, Goldblum \etal~\cite{goldblum2020adversarially} first proposed Adversarial Robustness Distillation (ARD). By distilling the robustness of the teacher, the student obtained higher robustness than AT from scratch. RSLAD~\cite{zi2021revisiting} found that using pseudo-labels provided by a robust teacher can further improve the robustness. IAD~\cite{zhu2021reliable} found that the guidance from the teacher model is progressively unreliable and proposed a multi-stage strategy to address this issue. However, these methods require access to the training set and the parameters of the teacher. BEST~\cite{li2023extracting} proposed to steal the robustness of the target model in the black-box setting, but it necessitated a proxy dataset. DFARD~\cite{wang2023model} proposed ARD in a data-free setting, yet still required white-box access.

\section{How To Steal Robustness?}\label{sec:robustness_steal}
In this section, for a fair comparison, we first assume that the attacker can obtain a proxy dataset for attacks as~\cite{li2023extracting}. Given a target model $M_{T}$ for a classification task that is built via AT and exhibits certain adversarial robustness, our goal is to train a clone model $M_{C}$ that performs similarly to $M_{T}$ on both clean and adversarial examples. The attacker has no prior knowledge of $M_{T}$, including the architecture, parameters, and training strategies, and is only granted black-box access to $M_{T}$. We consider the typical MLaaS scenario, where $M_{T}$ only returns top-1 predicted labels, \ie, hard-label setting. 
Most MSA methods usually use proxy data as query samples to query $M_T$, with the purpose of merely stealing accuracy. Recently, there have been new attempts of robustness stealing~\cite{li2023extracting}. Here, we provide a comprehensive summary and analysis.

\begin{table}
    \centering
    \renewcommand{\arraystretch}{1.13}
    \small
    \begin{tabular}{ccccc}
    \toprule
    Cloen Model                & Method                                                        & Clean Acc       & Robust Acc      & Avg.            \\ \hline
    \multirow{4}{*}{ResNet18}  & AT                                                            & 34.22           & \textbf{22.58}          & 28.40          \\
                               & UE                                                            & 43.98           & 15.78          & 29.88          \\
                               & AE                                                            & 34.76           & 20.32          & 27.54          \\
                               & \textbf{HEE} & \textbf{49.38}  & 19.60 & \textbf{34.49} \\ \hline
    \multirow{4}{*}{MobileNet} & AT                                                            & 34.42           & \textbf{22.64}          & 28.53          \\
                               & UE                                                            & 48.84           & 14.86          & 31.85          \\
                               & AE                                                            & 35.46           & 21.00          & 28.23          \\
                               & \textbf{HEE} & \textbf{50.12} & 18.98 & \textbf{34.55} \\ \bottomrule
    \end{tabular}
    \caption{Quantitative comparison of attacks using different query samples. $M_T$ is ResNet18 trained on CIFAR-10 training set, $M_C$ is ResNet18 or MobileNetV2, the proxy dataset is a random half of CIFAR-100 test set. Clean Acc and Robust Acc represent the accuracy (\%) of clean samples and adversarial samples generated by PGD, respectively.}
    \label{tab:query_sample_comp}
\end{table}

\subsection{Adversarial Training (AT)}
A straightforward way to steal robustness is to conduct AT during the attack. Specifically, the attacker first queries $M_{T}$ with query samples $x_q$ and obtains top-1 predictions $y_q$, and then conducts standard AT on $M_{C}$ with $(x_q,y_q)$ as follows:
\begin{linenomath*}
\begin{equation}\label{eq:AT}
\min _{\theta_{M_C}} \mathbb{E}_{(x_q, y_q) \in \mathcal{D}_p}(\underset{\delta}{\arg \max } \ \mathcal{L}_{\text{CE}}(\theta_{M_C}, x_q+\delta, y_q)) ,
\end{equation}
\end{linenomath*}
where $\mathcal{L}_{CE}$ is the cross-entropy loss function, $\theta_{M_C}$ represents parameters of the clone model, and $x_q + \delta$ represents the adversarial examples generated using PGD~\cite{madry2017towards}. However, this method will also inherit the shortcomings of AT, which will seriously reduce the clean accuracy.

\subsection{Uncertain Examples (UE)} 
In addition, Li~\etal introduced Uncertain Examples (UE) to achieve robustness stealing. The main idea is to find samples that the model predicts with the highest uncertainty across all classes and use them to query $M_T$. The construction process of UE is similar to that of AE, \ie, iteratively adds small noise to the query sample. Specifically, first assign the same target $Y=[1/K,\dots,1/K]$ ($K$ is the number of classes) to each query sample, 
then use the Kullback–Leibler (KL) divergence to compute and minimize the distance between the prediction of $M_T$ and Y.
Given a starting point $x_{\text{UE}}^{0}$ which is a random neighbor of the original query sample, an iterative update is performed with:
\begin{linenomath*}
\begin{equation}\label{eq:UE}
\begin{aligned}
x_{\text{UE}}^{t+1}=& \Pi_{\mathcal{B}_{\epsilon}[x_{\text{UE}}]}(x_{\text{UE}}^{t}-  \\
&\alpha \cdot \operatorname{sign}(\nabla_{x_{\text{UE}}^{t}}\mathcal{L}_{\text{KL}} (M_{C} (x_{\text{UE}}^{t} ) \| Y))) ,
\end{aligned}   
\end{equation}
\end{linenomath*}
where $\mathcal{L}_{\text{KL}}(\cdot \| \cdot)$ is the KL divergence, $\nabla_{x_{\text{UE}}^{(t)}}$ denotes the gradient of the loss function w.r.t. the uncertain example $x_{\text{UE}}^{t}$ in step $t$, $\alpha$ denotes step size, and $\Pi_{\mathcal{B}_{\epsilon}[x_{\text{UE}}]}(\cdot)$ projects its input onto the $\epsilon$-bounded neighborhood of the original one.

However, although this method alleviates the negative impact on clean accuracy, it reduces the robustness of $M_C$, as shown in~\Tref{tab:query_sample_comp}. We analyze the limitations of UE:

\begin{itemize}
    \item First, using the manually defined $Y$ as the target in~\Eref{eq:UE} is too hard for optimization, which will make UE more inclined to be located at the junction of the classification boundaries of all classes. The iterative process in~\Fref{fig:db_comp}(a) confirms this conjecture.
    \item Additionally, UE, like AE, uses the $l_{\infty}$-norm to constrain the magnitude of added noise with the aim of improving the visual quality of samples. However, this will sacrifice attack performance, as shown in \Tref{tab:HEE_ablation}. Actually, the constraint is not necessary in MSA, because the attacker typically uses abnormal samples or even unnatural synthetic samples in the data-free setting
\end{itemize}

\begin{table}
    \centering
    \renewcommand{\arraystretch}{1.13}
    \small
    \begin{tabular}{lccc}
    \toprule
                                  & Clean Acc & Robust Acc & Avg. \\ \hline
    HEE            & \textbf{50.12}            & \textbf{18.98}              & \textbf{34.55}        \\
    HEE \textbf{w/o} $\mathcal{H}(\cdot)$ & 48.26             & 18.26              & 33.26        \\
    HEE \textbf{w/} $l_\infty$-norm & 48.86             & 15.90              & 32.38        \\ \bottomrule
    \end{tabular}
    \caption{Ablation study about HEE. "w/o $\mathcal{H}(\cdot)$" represents the same objective as UE. "w/ $l_\infty$-norm" indicates that $l_\infty$-norm constraints are used in the process of constructing HEE like UE. $M_C$ is MobileNetV2.}
    \label{tab:HEE_ablation}
\end{table}

\begin{figure*}
\centering
\includegraphics[width=0.95 \linewidth]{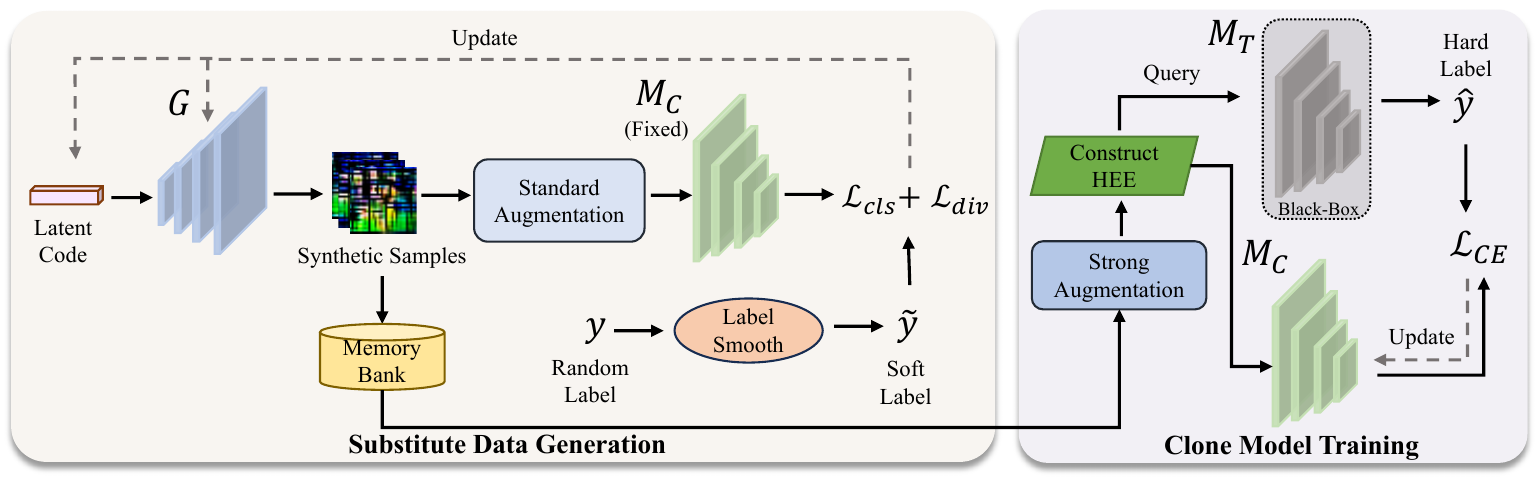}
\caption{The pipeline of our DFHL-RS attack, which consists of two alternately executed stages in each epoch. In the first stage, we optimize latent code and generators to generate substitute data and store it in the memory bank. In the second stage, we sample multiple batches from the memory bank and construct HEE to query the target model for hard labels, then use them to update the parameters of the clone model.}
\label{fig:framework}
\end{figure*}

\subsection{High-Entropy Examples (HEE)} 
We hope to construct samples that can characterize a more complete shape of the classification boundary, not just the junction as UE. Intuitively, samples located near the classification boundaries usually have larger prediction entropy, and the model will give similar confidence in multiple classes. Therefore, we propose the concept of High-Entropy Examples (HEE), which can be constructed by directly maximizing the prediction entropy as follows:
\begin{linenomath*}
\begin{equation}\label{eq:HEE}
\begin{aligned}
x_{\text{HEE}}^{t+1}=x_{\text{HEE}}^{t}+\alpha \cdot \operatorname{sign}(\nabla_{x_{\text{HEE}}^{t}} \mathcal{H}(M_{C}(x_{\text{HEE}}^{t}))),
\end{aligned}   
\end{equation}
\end{linenomath*}
where $\mathcal{H}(\cdot)$ calculates the entropy of the prediction, $\nabla_{x_{\text{HEE}}^{(t)}}$ denotes the gradient of the entropy loss function w.r.t. the high-entropy example $x_{\text{HEE}}^{t}$ in step $t$ and $\alpha$ is the step size. Note that we no longer use the $l_{\infty}$-norm to constrain the modification of the sample during iterations. 

Compared to manually assigning the same confidence value to all classes in UE, \Eref{eq:HEE} provides an adaptive optimization objective, allowing samples to explore among several similar classes rather than all classes. It can be seen from \Fref{fig:db_comp}(b) that HEE can gradually distribute near the classification boundaries over steps. Therefore, keeping the predictions of $M_C$ on these samples consistent with $M_T$ enables it to learn the shape of the classification boundary.

\subsection{Adversarial Examples (AE)} 
Previous work~\cite{he2018decision, li2023extracting} noticed that AE is also close to the classification boundaries, which may achieve the similar effect of UE and HEE. For robustness stealing, we can first obtain pseudo-labels of proxy data by querying $M_T$ and use the standard PGD to construct AE. Then we use AE to query $M_T$ again for corresponding labels to train $M_C$.

\subsection{Comparison of Different Query Samples} 
For quantitative evaluation, we follow the same setting as~\cite{li2023extracting}. As shown in~\Tref{tab:query_sample_comp}, although AT and AE can make $M_T$ obtain higher robustness, they will greatly reduce the clean accuracy. While UE can improve the clean accuracy, the robustness will be significantly reduced. Our proposed HEE achieves the best balance between clean accuracy and robustness. In \Tref{tab:HEE_ablation}, we also conduct an ablation study on different components of HEE.

To illustrate the superiority of HEE over UE in characterizing classification boundaries, we make a two-dimensional dataset with four classes to train a MLP (see appendix for details), as shown in \Fref{fig:db_comp}. As the iteration steps increase during the construction process, UE will gradually concentrate at the junction of classification boundaries, while HEE will be uniformly distributed around the classification boundaries, thus characterizing its more complete shape. This is in line with our previous analysis. Therefore, querying with HEE allows $M_C$ to better approximate the classification boundaries of $M_T$.

\section{Data-Free Hard-Label Robustness Stealing}
In this section, we further explore the challenging data-free scenario, where the attacker CAN NOT obtain any similar natural samples as proxy data.
The framework is illustrated in \Fref{fig:framework} and the training algorithm is in the appendix.

\subsection{Overview}
 Unlike previous work~\cite{sanyal2022towards,beetham2023dual}, which uses $M_T$ as a discriminator and plays a min-max game, we decouple the training process of the generator and the training process of $M_C$ into two stages: 1) Substitute Data Generation and 2) Clone Model Training. In the first stage, we train a generator to synthesize substitute data to approximate the distribution of the target data and store them in a memory bank. Due to the hard-label setting, we use $M_C$ to guide the training of the generator, rather than $M_T$ as in previous work~\cite{kariyappa2021maze, truong2021data}. In the second stage, we randomly sample multiple batches of substitute data from the memory bank and use \Eref{eq:HEE} to construct HEE, then use HEE to query $M_T$ for hard labels to optimize the parameters of $M_C$. 
 These two stages are executed alternately in each epoch.

\subsection{Substitute Data Generation} \label{sec:first_stage}
We adopt the ``batch-by-batch" manner to synthesize substitute data, that is, only one batch of images will be synthesized by a new generator in each epoch. Specifically, at the beginning of an arbitrary epoch $i$, we resample a batch of latent code $z_i \sim \mathcal{N}(0,1)$ and reinitialize the generator $G_{i}$ with parameters from the last epoch. Then we randomly sample a batch of corresponding labels $\tilde{y}_i$ from the uniform distribution. Intuitively, we hope that the images synthesized by the generator can be classified into the specified class by $M_T$. This means that the distribution of the synthetic data is similar to the target data. For this, we can iteratively optimize the latent code $z_i$ as well as the parameters $\theta_{G_i}$ as follows:
\begin{linenomath*}
\begin{equation}\label{eq:cls_loss_1}
\begin{aligned}
\mathcal{L}_{\text{cls}} = \underset{z_i, \theta_{G_i} }{\arg\min}  \ \mathcal{L}_{\text{CE}} (M_T(G_i(z_i;\theta_{G_i}), \tilde{y}_i)),
\end{aligned}   
\end{equation}
\end{linenomath*}

However, the backpropagation of \Eref{eq:cls_loss_1} will violate the principles of a black-box setting. The gradient estimation techniques~\cite{truong2021data,kariyappa2021maze} also cannot be applied due to the unavailability of soft labels. To address this issue, we use the latest $M_C$ as a surrogate for $M_T$, providing gradients for optimization. The optimization problem can be formulated as follows:
\begin{linenomath*}
\begin{equation}\label{eq:cls_loss_2}
\begin{aligned}
\mathcal{L}_{\text{cls}} = \underset{z_i, \theta_{G_i} }{\arg\min} \ \mathcal{L}_{\text{CE}} (M_{C_{i-1}}(G_i(z_i;\theta_{G_i}), \tilde{y}_i)) .
\end{aligned}   
\end{equation}
\end{linenomath*}

To prevent overfitting of synthetic data to the clone model, we use a small number of iterations (see appendix for details). We perform standard data augmentation on synthetic samples to ensure that they are not deceptive (\eg, adversarial example). In addition, we also adopt label smoothing~\cite{szegedy2016rethinking} to soften the corresponding labels to further alleviate overfitting.

Moreover, merely ensuring the generator produces images from the desired distribution is inadequate, as it may still encounter issues such as mode collapse and insufficient diversity~\cite{chen2019data, sanyal2022towards}. In a hard-label setting, a lack of diversity among classes can significantly hamper the learning process of $M_C$, particularly for the less represented classes. 
Therefore, to promote the generation of diverse images across all classes in each batch, we adopt a class-diversity loss as follows:
\begin{linenomath*}
\begin{equation}\label{eq:div_loss}
\begin{aligned}
\mathcal{L}_{\text{div}}=\sum_{j=0}^K \alpha_j \log \alpha_j, \alpha_j=\frac{1}{N} \sum_{i=1}^N \operatorname{softmax}(M_{C}\left(x_i\right))_j,
\end{aligned}   
\end{equation}
\end{linenomath*}
where $K$ denotes the number of classes, $\alpha_j$ denotes the expected confidence value for every class $j$ over a batch with $N$ samples and the $\mathcal{L}_\text{div}$ calculates the negative entropy. When the loss is minimized, the entropy of the number of synthetic samples per class gets maximized such that each class has a similar amount of samples.

By combining the aforementioned two loss functions, we obtain the final objective:
\begin{linenomath*}
\begin{equation}\label{eq:final_loss}
\begin{aligned}
\mathcal{L}_{\text{gen}} = \mathcal{L}_{\text{cls}} + \lambda \cdot \mathcal{L}_{\text{div}},
\end{aligned}   
\end{equation}
\end{linenomath*}
where $\lambda$ is a hyperparameter for balancing two different terms. Note that we does not require training a global generator to model the entire distribution of the target data. Instead, in each epoch, only a temporary generator is trained to handle a specific data distribution for one batch. Besides, we keep the parameters of $M_C$ fixed in this stage. After optimization by \Eref{eq:final_loss}, we store this batch of synthetic data into the memory bank and discard the generator.

\begin{table*}[ht]
    \centering
    \small
    \renewcommand{\arraystretch}{1.13}
    \begin{tabular}{ccccccccc}
    \toprule
    Target   Data             & Data-Free                   & Method             & Clean Acc      & FGSM           & PGD-20         & PGD-100        & CW-100         & AA     \\ \hline
    \multirow{6}{*}{CIFAR10}  & /                           & Target Model       & \underline{82.57} & \underline{56.99} & \underline{51.31} & \underline{50.92} & \underline{49.68} & \underline{47.91} \\ \cline{2-9} 
                              & $\times$                    & BEST~\cite{li2023extracting}      & 67.31         & 32.68         & 27.48          & 27.27        & 28.33        & 28.33        \\ \cline{2-9} 
                              & \multirow{4}{*}{\checkmark} & Data-Free AT                 & 36.15          & 15.86          & 11.87          & 11.73          & 12.03          & 11.43          \\
                              &                             & Data-Free AE                 & 67.78          & 32.20          & 28.20          & 28.07          & 28.50          & 27.89          \\
                              &                             & Data-Free UE                 & 74.24          & 39.78          & 35.00          & 34.80          & 35.28          & 34.41          \\
                              &                             & \textbf{DFHL-RS(Ours)} & \textbf{77.86} & \textbf{44.94} & \textbf{40.07} & \textbf{39.87} & \textbf{40.64} & \textbf{39.51} \\ \hline
    \multirow{6}{*}{CIFAR100} & /                           & Target Model       & \underline{56.76} & \underline{31.96} & \underline{28.96} & \underline{28.83} & \underline{26.84} & \underline{26.84} \\ \cline{2-9} 
                              & $\times$                    & BEST~\cite{li2023extracting}               & 28.33          & 18.78          & 18.78          & 15.08          & 16.28          & 14.59          \\ \cline{2-9} 
                              & \multirow{4}{*}{\checkmark} & Data-Free AT                 & 20.22          & 9.79           & 8.63           & 8.62           & 8.73           & 8.74           \\
                              &                             & Data-Free AE                 & 37.76          & 15.41          & 13.15          & 13.00          & 13.90          & 12.77          \\
                              &                             & Data-Free UE        & 39.25          & 16.88          & 14.18          & 14.06          & 14.94          & 14.94          \\
                              &                             & \textbf{DFHL-RS(Ours)} & \textbf{51.94} & \textbf{23.68} & \textbf{20.02} & \textbf{19.88} & \textbf{20.91} & \textbf{19.30} \\ \bottomrule
    \end{tabular}
     \caption{Attack performance comparison between different methods. $M_T$ and $M_C$ are ResNet18. The AT strategy is PGD-AT. We report the clean accuracy and robust accuracy (\%) of $M_C$.}
    \label{tab:baseline_comp}
\end{table*}

\begin{table}
    \centering
    \renewcommand{\arraystretch}{1.13}
    \small
    \begin{tabular}{ccccc}
    \toprule
    \begin{tabular}[c]{@{}c@{}}Target\\ Model\end{tabular} & AT Strategy & Clean Acc & PGD-100 & AA    \\ \hline
    \multirow{3}{*}{ResNet18}                              & PGD-AT          & \textbf{77.86}     & \textbf{39.87}   & \textbf{39.51} \\
                                                           & TRADES      & 72.15     & 37.24   & 37.09 \\
                                                           & STAT-AWP    & 72.24     & 37.79   & 37.73 \\ \hline
    \multirow{3}{*}{WideResNet}                            & PGD-AT          & \textbf{77.75}     & 36.65   & 36.55 \\
                                                           & TRADES      & 71.04     & 34.20   & 33.94 \\
                                                           & STAT-AWP    & 73.81     & \textbf{38.16}   & \textbf{38.13} \\ \bottomrule
    \end{tabular}
    \caption{Attack performance under different architectures of $M_T$ and various AT strategies. $M_C$ is ResNet18.}
    \label{tab:different_target}
\end{table}

\subsection{Clone Model Training} 
In the second stage, we train $M_C$ with synthetic substitute data. However, if we optimize $M_C$ using only one batch of samples synthesized in the first stage of the current epoch, it will suffer from catastrophic forgetting~\cite{binici2022preventing, binici2022robust,do2022momentum}. The reason is that the substitute data is synthesized under the guidance of $M_C$, as shown in \Eref{eq:cls_loss_2}. With each epoch, the gap between $M_C$ and $M_T$ decreases, causing a shift in the distribution of synthetic data over time. Hence, if $M_C$ does not periodically relearn previously synthesized samples, the knowledge acquired during the early training phase may be lost. This can result in performance degradation or even failure to converge over time.

To address this issue, we use a memory bank to store all previously synthesized samples in the first stage. In each epoch, we optimize $M_C$ for $N_C$ steps. In each step, we randomly select a batch of synthetic samples from the memory bank. Then we use \Eref{eq:HEE} to construct HEE $x_{\text{HEE}}$, but before that, we need to perform strong augmentation (see appendix for details) on $x$ to promote the diversity . This is beneficial to characterize more complete classification boundaries. Finally, we feed $x_{\text{HEE}}$ into $M_T$ to query the hard label $\hat{y}$ and use cross-entropy loss to optimize $M_C$ as follows:
\begin{linenomath*}
\begin{equation}\label{eq:training_loss}
\begin{aligned}
\mathcal{L}_{C} = \mathcal{L}_{\text{CE}} (M_C(x_{\text{HEE}}), \hat{y}).
\end{aligned}   
\end{equation}
\end{linenomath*}
By minimizing $\mathcal{L}_{C}$, $M_C$ can obtain similar classification boundaries as $M_T$ by imitating its predictions for $x_{\text{HEE}}$, thus simultaneously stealing the accuracy and robustness.

\section{Experiments}\label{sec:experiment}
In this section, we first provide the detailed experimental settings. To demonstrate the effectiveness of our method, we evaluate it from several perspectives.

\subsection{Experimental Settings}

\paragraph{Datasets.}
We consider two benchmark datasets commonly used in AT research~\cite{madry2017towards, zhang2019theoretically, li2023squeeze}, CIFAR-10 and CIFAR-100~\cite{krizhevsky2009learning}, as target datasets. Prior work requires a proxy dataset with samples from the same distribution~\cite{tramer2016stealing,jagielski2020high,orekondy2019knockoff} or the same task domain~\cite{sanyal2022towards,li2023extracting} as the target dataset. However, we avoid using any natural data.

\paragraph{Models.} We evaluate our attack on different models with various architectures. $M_T$ is selected from ResNet18~\cite{he2016deep} and WideResNet-34-10~\cite{zagoruyko2016wide}. $M_C$ may be different from $M_T$ in architecture, so we use two additional models, \ie, ResNet34 and MobileNetV2~\cite{sandler2018mobilenetv2}. We employ two commonly used AT strategies, namely PGD-AT~\cite{madry2017towards} and TRADES~\cite{zhang2019theoretically}, and a state-of-the-art method, STAT-AWP~\cite{li2023squeeze}, to improve the robustness of $M_T$. We follow the same generator architecture as~\cite{ijcai2021p327,fang2022up}.

\paragraph{Baselines.}
This is the first work to achieve data-free hard-label robustness stealing attack. The closest work to ours is \textit{BEST}~\cite{li2023extracting}, but it requires a proxy dataset to attack. Therefore, we also make some modifications to the second stage of our framework as baselines: (1) \textit{Data-Free AT}: use synthetic samples to query $M_T$ for corresponding labels and then perform AT on $M_C$. (2) \textit{Data-Free UE}: construct uncertain examples with synthetic samples to query $M_T$. (3) \textit{Data-Free AE}: construct adversarial examples with synthetic samples to query $M_T$.

\paragraph{Metrics.}
In this task, $M_C$ should not only have good usability, but also need to have certain robustness. Therefore, in addition to considering clean accuracy, \ie, the accuracy over clean samples, we also measure robust accuracy against various adversarial attacks, including FGSM~\cite{goodfellow2014explaining}, PGD-20, PGD-100~\cite{madry2017towards}, CW-100~\cite{carlini2017towards} and AutoAttack (AA)~\cite{croce2020reliable}. The attack settings are $\epsilon = 8/255$ and $\eta = 2/255$. The number of attack step is 20 for PGD-20, and 100 for PGD-100 and CW-100.

\paragraph{Implementation Details.}
For substitute data generation, we use Adam optimizer with $\beta=(0.5,0.999)$ and set the hyperparameter $\lambda=3$ in \Eref{eq:final_loss}. For CIFAR-10, we set learning rates $\eta_{G}=0.002, \eta_{z}=0.01$, number of iterations $N_G = 10$ and the label smoothing factor is set to $0.2$. For CIFAR-100, we set learning rates $\eta_{G}=0.005, \eta_{z}=0.015$, number of iterations $N_G = 15$ and the label smoothing factor is set to $0.02$.
For training $M_C$, we use SGD optimizer with an initial learning rate of $0.1$, a momentum of $0.9$ and a weight decay of $1e-4$. For constructing HEE, the step size $\alpha$ in \Eref{eq:HEE} is set to $0.03$ and the number of iterations is set to $10$. We set the iterations of the clone model $N_C = 500$. The batch sizes for CIFAR-10 and CIFAR100 are set to $B=256$ and $B=512$, respectively. We apply a cosine decay learning rate schedule and the training epoch is $E=300$.

\begin{table}
    \centering
    \renewcommand{\arraystretch}{1.13}
    \small
    \begin{tabular}{cccc}
    \toprule
    Clone Model & Clean Acc & PGD-100 & AA    \\ \hline
    MobileNet   & 73.50     & 33.32   & 33.19 \\
    ResNet34    & \textbf{78.49}     & 40.25   & 39.97 \\
    WideResNet  & 77.38     & \textbf{40.66}   & \textbf{40.33} \\ \bottomrule
    \end{tabular}
     \caption{Attack performance using different architectures of $M_C$. $M_T$ is ResNet18 using PGD-AT strategy.}
    \label{tab:different_clone}
\end{table}

\subsection{Experimental Results}

\paragraph{Performance on Robustness Stealing.}

We first compare the attack effects of different methods by evaluating the clone model, as shown in \Tref{tab:baseline_comp}. Our DFHL-RS significantly outperforms the baselines in both accuracy and robustness. When the target data is CIFAR-10, our method achieves 77.86\% clean accuracy, which is only 4.71\% lower than the target model. Meanwhile, it also achieves 39.51\% robust accuracy against AA, which is only 8.40\% lower than the target model. When the target data is CIFAR100, our method achieves 51.94\% clean accuracy and 19.30\% robust accuracy against AA, which are only 4.82\% and 7.53\% lower than the target model, respectively. It should be emphasized that our method does not require any natural data, but still outperforms BEST which requires a proxy dataset.

\begin{table*}[ht]
    \centering
    \small
    \renewcommand{\arraystretch}{1.13}
    \begin{tabular}{cccccccc}
    \toprule
    \multirow{2}{*}{Epsilon($\epsilon$)} & \multirow{2}{*}{Attack} & \multirow{2}{*}{\textit{White-box}} & \multirow{2}{*}{\textit{AT Proxy}} & \multicolumn{4}{c}{Data-Free Model   Stealing Method} \\ \cline{5-8} 
                             &                         &                                     &                                    & MAZE    & DFME   & DFMS-HL  & \textbf{DFHL-RS(Ours)}  \\ \hline
    \multirow{2}{*}{$4 / 255$}   & FGSM                    & \textit{12.49}                      & \textit{9.84}                      & 4.36    & 3.90   & 6.09     & \textbf{10.29}          \\
                             & PGD-20                     & \textit{13.54}                      & \textit{10.27}                     & 4.31    & 3.71   & 6.11     & \textbf{10.54}          \\ \hline
    \multirow{2}{*}{$8 / 255$}   & FGSM                    & \textit{25.58}                      & \textit{21.44}                     & 10.51   & 9.25   & 13.95    & \textbf{22.77}          \\
                             & PGD-20                     & \textit{31.25}                      & \textit{24.33}                     & 10.13   & 9.06   & 14.52    & \textbf{24.59}          \\ \hline
    \multirow{2}{*}{$12 / 255$}  & FGSM                    & \textit{36.99}                      & \textit{31.94}                     & 17.42   & 15.57  & 22.07    & \textbf{34.30}          \\
                             & PGD-20                     & \textit{49.02}                      & \textit{40.44}                     & 71.73   & 15.80  & 23.93    & \textbf{39.25}          \\ \bottomrule
    \end{tabular}
    \caption{Attack Success Rate (ASR) (\%) of transfer-based adversarial attack using the clone model obtained by different methods as a surrogate model. AT Proxy represents the adversarial training model on the target data. The target model is trained on CIFAR-10 dataset using PGD-AT with $\epsilon=8/255$. }
    \label{tab:transfer_attack}
\end{table*}

\begin{table}
\centering
\renewcommand{\arraystretch}{1.13}
\small
   \begin{tabular}{ccccc}
    \toprule
    \multicolumn{2}{c}{Modification}    & \begin{tabular}[c]{@{}c@{}}Query\\ Budget\end{tabular} & Clean Acc & AA \\ \hline
    \multicolumn{2}{c}{Default}         & 38.4M                                                  & 77.86     & 39.51      \\ \hline
    B      & $256 \rightarrow 128$    & \multirow{3}{*}{19.2M}                                 & \textbf{75.79}     & \textbf{36.42 }     \\
    E      & $300  \rightarrow 150$   &                                                        & 73.81     & 34.73      \\
    $N_C$    & $500 \rightarrow 250$    &                                                        & 74.07     & 35.19      \\ \bottomrule
    \end{tabular}
    \caption{Attack performance after modifying hyperparameters to halve query budget.}
    \label{tab:query_budget}
\end{table}

\paragraph{Different Model Architectures and AT Strategies.}
In real scenarios, MLaaS providers may use different architectures and strategies to improve the robustness of the target model, so we study the impact of different target model architectures using various AT strategies on the attack. See appendix for the performance of all target models. As shown in \Tref{tab:different_target}, when the target model is ResNet18, PGD-AT is more vulnerable to robustness stealing attacks. Although the target model using PGD-AT has lower robustness than using STAT-AWP, the clone model obtained by the attack has the best performance. When the target model is WideResNet with a different architecture, STAT-AWP can make the clone model obtain the highest robustness, but PGD-AT still has the best clean accuracy. Consider the black-box setting, the attacker may use different clone model architectures for the attack. As shown in \Tref{tab:different_clone}, close attack performance is achieved when the clone model is ResNet34 and WideResNet, while the attack performance drops slightly when the clone model is MobileNetv2, probably due to the larger difference in architecture. In conclusion, our method achieves stable attack performance across different configurations.

\paragraph{Transfer-Based Adversarial Attacks.}

We study the effectiveness of black-box transfer-based adversarial attacks on the target model using the clone model as a surrogate. As shown in \Tref{tab:transfer_attack}, although black-box attacks on robust models are extremely challenging~\cite{dong2019evading,Dong_2020_CVPR}, our method significantly improves the attack success rate (ASR) at various $\epsilon$ compared to the baselines. In most cases, our method even slightly exceeds the AT Proxy, which directly uses the target data to train a surrogate model, and is slightly worse than the white-box attack. As $\epsilon$ decreases, the gap between our method and the white-box attack gets smaller.

\paragraph{Attack Cost of DFHL-RS.}
We also consider the attack cost of our DFHL-RS. In terms of query budget, our method eliminates the need for any querying during the data generation stage. It only requires querying the target model with each HEE sample for the corresponding label. Therefore, the total query budget depends on the training epoch, batch size and the clone model iterations, \ie, $E \times B \times N_C$. Our default query budget for CIFAR-10 is 38M, whereas it is 20M for DFME and DS, and 30M for MAZE. This is reasonable because acquiring robustness typically comes at a higher cost. By modifying these three hyperparameters, we can control the query budget as shown in \Tref{tab:query_budget}. When the query budget is halved, the clean accuracy and robust accuracy are only reduced by approximately 2\% and 3\%, respectively. More results can be found in the appendix.

In terms of time overhead, since only a few iterations per epoch are sufficient for the first stage, it takes very little time. The time overhead of the second stage is mainly concentrated on the inner loop for constructing HEE, and its computational complexity is similar to that of standard AT. The time overhead mainly depends on the number of training samples per epoch, \ie, $B \times N_C$, which is comparable to many AT techniques.

\paragraph{Ablation Study.}
We further investigate the effects of the various components in our DFHL-RS framework as shown in \Tref{tab:DFHL_RS_ablation}. When the diversity loss in \Eref{eq:final_loss} is discarded, the clean accuracy and robust accuracy both decrease. In addition, removing the label smoothing and standard augmentation in the first stage will also affect the attack performance. The strong data augmentation in the second stage improves the attack effect by promoting the diversity of HEE.

\begin{table}
\centering
\renewcommand{\arraystretch}{1.13}
\small
    \begin{tabular}{ccc}
    \toprule
                              & Clean Acc      & AA     \\ \hline
    DFHL-RS                   & \textbf{77.86} & \textbf{39.51} \\
    \textbf{w/o} Div Loss              & 76.87          & 38.87          \\
    \textbf{w/o} Label Smoothing       & 77.10          & 38.42          \\
    \textbf{w/o} Standard Augmentation & 74.99          & 36.80          \\
    \textbf{w/o} Strong Augmentation   & 77.13          & 38.66          \\ \bottomrule
    \end{tabular}
    \caption{Ablation study of different components.}
    \label{tab:DFHL_RS_ablation}
\end{table}

\section{Conclusion} 
In this paper, we first explore a novel and challenging task called Data-Free Hard-Label Robustness Stealing (DFHL-RS) attack, which can steal both clean accuracy and adversarial robustness by simply querying a target model for hard labels without the participation of any natural data. Experiments show that our method achieves excellent and stable attack performance under various configurations. Our work aims to advance research to improve security and privacy by raising awareness of the vulnerabilities of machine learning models through attacks.

\section*{Acknowledgements}
This work was supported in part by the Natural Science Foundation of China under Grant  U20B2047, 62102386, U2336206, 62072421 and 62121002, and by Xiaomi Young Scholars Program.
\bibliography{main} 

\clearpage

\appendix

\section{Appendix}

\subsection{Details of Figure 1}

To illustrate the superiority of HEE over UE in characterizing classification boundaries, 
we make a two-dimensional toy dataset comprising of four classes. Each class is generated by sampling data points from a two-dimensional Gaussian distribution. 
The means of the Gaussian distributions for the four classes are: (0, 12), (0, -12), (12, 0), (-12, 0). The corresponding standard deviations are: (5, 0.5), (5, 0.5), (0.5, 5), (0.5, 5). In addition, we use a MLP which consists of a linear layer with output size of 10 followed by a ReLU, and a final linear layer with output size of 4 for classification. We provide pytorch-style pseudocodes in Pseudocode 1 and Pseudocode 2. We train the MLP using the SGD optimizer with a learning rate of 0.02 for 100 epochs. For constructing UE and HEE, we discard the $l_{\infty}$-norm constraint and use a step size of 1, the number of iterations is set from 1 to 10.
\label{toy_dataset}
\begin{lstlisting}[title={Pseudocode 1: Make a toy dataset.}]
# n is the number of samples per class
def toy_dataset(n=100):
    x = torch.zeros(n * 4, 2)
    y = torch.zeros(n * 4, dtype=torch.long)
    means = [[0, 12],
             [0, -12],
             [12, 0],
             [-12, 0]]
    stds = [[5, 0.5],
            [5, 0.5],
            [0.5, 5],
            [0.5, 5]]
            
    for i, (mean, std) in enumerate(zip(means, stds)):
        x[i * n: (i + 1) * n] = torch.randn(n, 2) * torch.tensor(std) + torch.tensor(mean)
        y[i * n: (i + 1) * n] = i
    
    return x, y
\end{lstlisting}

\label{MLP}
\begin{lstlisting}[title={Pseudocode 2: MLP Architecture.}]
class MLP(torch.nn.Module):
    def __init__(self, n_features=2, n_hidden=10, n_output=4):
        super(MLP, self).__init__()
        self.hidden = torch.nn.Linear(n_features, n_hidden)
        self.predict = torch.nn.Linear(n_hidden, n_output)

    def forward(self, x):
        x = F.relu(self.hidden(x))
        x = self.predict(x)
        return x
\end{lstlisting}

\subsection{Algorithm}
The training process of our DFHL-RS framework is demonstrated in \Aref{algorithm_1}.

\begin{algorithm}
\SetAlgoLined
\KwInput{Target model $M_T$, generator $G(\cdot;\theta_{G})$, epochs $E$, generator iterations $N_G$ in each epoch, clone model iterations $N_C$ in each epoch, learning rate of generator $\eta_{G}$, learning rate of latent code $\eta_{z}, $learning rate of clone model $\eta_{C}$, memory bank $M_B$}

\KwOutput{Clone model $M_{C}(\cdot;\theta_{C})$}
    \For{$e = 1, \dots, E$}{
        \textit{// Substitute Data Generation} \\
        Initialize the generator $G$ \\
        Sample a batch of latent code $z \sim \mathcal{N}(0,1)$ and corresponding labels $\tilde{y}$  \\
        \For{$i = 1, \dots, N_{G}$} {
            Generate synthetic samples by $G(z;\theta_{G})$ \\
            Perform standard augmentation and label smoothing  \\
             Compute $\mathcal{L}_{\text{gen}}$ by Eq. (7) \\
            Update $ z \leftarrow \eta_{z} \nabla_{z} \mathcal{L}_{\text{gen}}$ \\
            Update $\theta_{G} \leftarrow \eta_{G} \nabla_{\theta_{G}} \mathcal{L}_{\text{gen}}$ \\
        }
        Save this batch of synthetic samples to $M_B$ \\
        \textit{// Clone Model Training} \\
        \For{$i = 1, \dots, N_{C}$} {
           Randomly sample a batch from $M_B$ and perform strong augmentation \\
           Construct HEE $x_{\text{hee}}$ by Eq. (3) \\
           Query $M_T$ with HEE for hard labels $\hat{y}$ \\
           Compute $\mathcal{L}_{C}$ by Eq. (8) \\
           Update $\theta_C \leftarrow \eta_{C} \nabla_{\theta_C}\mathcal{L}_{C}$ \\
        }
    }
 \caption{Training Process of DFHL-RS}
 \label{algorithm_1}
\end{algorithm}

\begin{table}[!h]
\centering
\renewcommand{\arraystretch}{1.15}
\small
   \begin{tabular}{cccccc}
    \toprule
    $N_G$     & 3     & 5     & 10             & 15    & 20    \\ \hline
    Clean Acc & 76.98 & 76.10 & \textbf{77.86} & 75.39 & 76.81 \\
    PGD-20    & 38.01 & 38.11 & \textbf{40.07} & 36.86 & 38.02 \\
    AA        & 37.58 & 37.72 & \textbf{39.51} & 36.58 & 37.56 \\ \bottomrule
    \end{tabular}
    \caption{Clean accuracy and robust accuracy (\%) of the clone model using different $N_G$ in the first stage.}
    \label{tab:different_NG}
\end{table}

\begin{table*}[!ht]
    \centering
    \small
    \renewcommand{\arraystretch}{1.15}
        \begin{tabular}{cccccccc}
        \toprule
        Target Model                & AT Strategy & Clean Acc & FGSM  & PGD-20 & PGD-100 & CW-100 & AA    \\ \hline
        \multirow{3}{*}{ResNet18}   & PGD-AT      & 82.57     & 56.99 & 51.31  & 50.92   & 49.68  & 47.91 \\
                                    & TRADES      & 80.85     & 55.92 & 51.23  & 51.18   & 48.09  & 48.09 \\
                                    & STAT-AWP    & 83.03     & 60.21 & 56.37  & 56.22   & 52.61  & 52.19 \\ \hline
        \multirow{3}{*}{WideResNet} & PGD-AT      & 86.48     & 61.16 & 54.90  & 54.38   & 54.12  & 52.14 \\
                                    & TRADES      & 84.38     & 60.50 & 54.65  & 54.40   & 53.21  & 52.21 \\
                                    & STAT-AWP    & 86.41     & 64.83 & 60.18  & 60.04   & 56.63  & 56.63 \\ \bottomrule
        \end{tabular}
    
     \caption{Clean accuracy and robust accuracy (\%) of target models with different architectures and  AT Strategies.}
    \label{tab:target_model_performance}
\end{table*}

\begin{figure*}[!t]
\centering
\includegraphics[width=0.88 \linewidth]{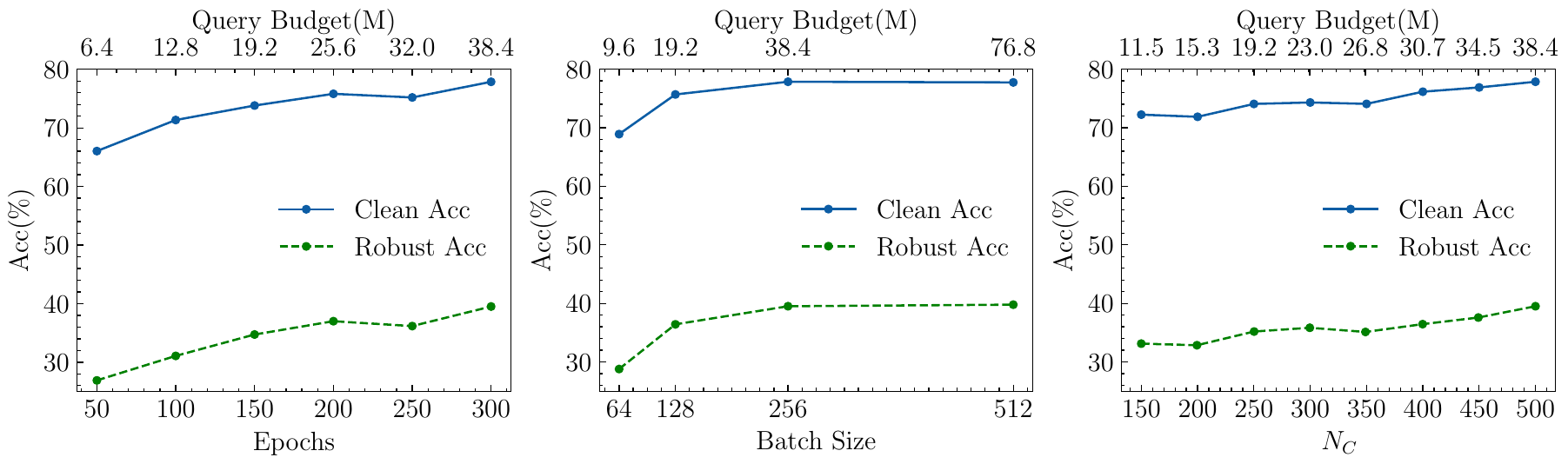}
\caption{Ablation study of query budgets corresponding to different configurations. We report the clean accuracy and robust accuracy against AA.}
\label{fig:query_budget}
\end{figure*}

\begin{table*}[!ht]
    \small
    \renewcommand{\arraystretch}{1.15}
    \resizebox{\linewidth}{!}{
    \begin{tabular}{cccccc}
            \toprule
            
            \multirow{3}{*}{\textbf{Approach}}  & \multicolumn{3}{c}{\textbf{Accuracy Extraction}}                                                                                                                                                                                        & \multicolumn{2}{c}{\textbf{Robustness Extraction}}                                                                                                                              \\ \cline{2-6} 
                                                & \multirow{2}{*}{ \small \textbf{White-Box}}                               & \multicolumn{2}{c}{\small \textbf{Black-Box}}                                                                                                                            & \multirow{2}{*}{\small \textbf{White-Box}}                                                 & \multirow{2}{*}{\small \textbf{Black-Box}}                                                     \\ \cline{3-4}
                                                &                                                                     & \small \textbf{Soft Label}                                                                  & \small \textbf{Hard Label}                                                        &                                                                                       &                                                                                         \\ \hline
            \multirow{2}{*}{\textbf{Data}}      & \multirow{2}{*}{Standard KD}                                        & \begin{tabular}[c]{@{}c@{}}KnockoffNets\\ \cite{orekondy2019knockoff}\end{tabular}   & \multirow{2}{*}{-}                                                         & \begin{tabular}[c]{@{}c@{}}ARD\\ \cite{goldblum2020adversarially}\end{tabular}        & \multirow{2}{*}{\begin{tabular}[c]{@{}c@{}}BEST\\ \cite{li2023extracting}\end{tabular}} \\
                                                &                                                                     & \begin{tabular}[c]{@{}c@{}}Black-box   dissector\\ \cite{wang2022black}\end{tabular} &                                                                            & \begin{tabular}[c]{@{}c@{}}RSLAD\\ \cite{zi2021revisiting}\end{tabular}               &                                                                                         \\ \hline
            \multirow{2}{*}{\textbf{Data-Free}} & \begin{tabular}[c]{@{}c@{}}ZSKD\\ \cite{nayak2019zero}\end{tabular} & \begin{tabular}[c]{@{}c@{}}MAZE\\ \cite{kariyappa2021maze}\end{tabular}              & \begin{tabular}[c]{@{}c@{}}DFMS-HL\\ \cite{sanyal2022towards}\end{tabular} & \multirow{2}{*}{\begin{tabular}[c]{@{}c@{}}DFARD\\ \cite{wang2023model}\end{tabular}} & \multirow{2}{*}{\textbf{\begin{tabular}[c]{@{}c@{}}DFHL-RS\\ (Ours)\end{tabular}}}      \\
                                                & \begin{tabular}[c]{@{}c@{}}DAFL\\ \cite{chen2019data}\end{tabular}  & \begin{tabular}[c]{@{}c@{}}DFME\\ \cite{truong2021data}\end{tabular}                 & \begin{tabular}[c]{@{}c@{}}DS\\ \cite{beetham2023dual}\end{tabular}        &                                                                                       &                                                                                         \\ \bottomrule
            \end{tabular}
        }
\caption{Taxonomy of prior works. } 
\label{tab: prior_works}
\end{table*}

\subsection{The Impact of Different $N_G$}
We modify the number of iterations $N_G$ in the first stage, the results are shown in \Tref{tab:different_NG}. We can observe that too large or too small $N_G$ cannot obtain the optimal attack performance. We speculate that a small $N_G$ leads to poor synthetic data due to insufficient optimization. However, a large $N_G$ leads to overfitting, making the synthetic data unsuitable for future training. Therefore, based on empirical experiments, we set $N_G=10$ for our attack.

\subsection{Performance of Target Models}
In our experiments, we train several target models with different architectures and various adversarial training strategies on CIFAR-10 dataset. As shown in \Tref{tab:target_model_performance},  we report the clean accuracy and robust accuracy of each target model.

\subsection{More Ablation Results of Query Budget}
As mentioned in the main manuscript, the query budget of our method depends on the training epoch $E$, batch size $B$ and clone model iterations $N_C$, i.e., $E \times B \times N_C$. 
In the default configuration, the three hyperparameters are $E=300, B=256, N_C=500$. By modifying them, we can control the corresponding query budget, as shown in \Fref{fig:query_budget}. The reduction of epoch has a greater impact on the attack performance, while the attack performance is relatively stable when $N_C$ is reduced. When the batch size is halved, there is only a slight decrease in attack performance, while doubling the batch size brings more query budget, but almost no improvement. In summary, our attack maintains stability and effectiveness, even with a limited query budget.

\subsection{Taxonomy of Prior Work}
We summarize existing work with different goals and varied levels of access to the target model as shown \Tref{tab: prior_works}. Our DFHL-RS is the first to achieve robust extraction in the data-free and black-box (i.e., hard-label) setting.

\subsection{Details of Data Augmentation}
In the first stage, we use the standard data augmentation, \ie, random cropping and random horizontal flipping, on the synthetic samples. This is commonly used for standard training CIFAR-10 classification models. In the second stage, we apply a strong data augmentation, including central cropping, adding Gaussian noise, random horizontal or vertical flipping, and random rotation, on the samples before constructing HEE.

\end{document}